\documentclass[sigconf,natbib]{acmart}

\usepackage{makecell}
\usepackage{multirow}
\usepackage{subcaption}
\usepackage{graphicx}
\usepackage{url}
\usepackage{enumitem}
\usepackage{array}
\usepackage{booktabs}
\usepackage{amsmath}
\usepackage{pifont}
\usepackage{tabularx}
\usepackage{comment}
\usepackage{hyphenat}
\usepackage{seqsplit}

\newcommand{\boldheading}[1]{%
    \vspace{0.5em} 
    \noindent\textbf{#1}\hspace{0.1em} 
}

\AtBeginDocument{%
  \providecommand\BibTeX{{%
    \normalfont B\kern-0.5em{\scshape i\kern-0.25em b}\kern-0.8em\TeX}}}

\setcopyright{acmcopyright}
\copyrightyear{2024}
\acmYear{2024}
\acmDOI{tbd}

\acmConference[Conf]{Conference}{Month DD, YYYY}{City, State}

\newcommand{\prompt}[1]{\textit{#1}}
\DeclareMathOperator*{\argmax}{arg\,max}

\begin{document}

\title{Imagery as Inquiry: Exploring a Multimodal Dataset for Conversational Recommendation}


\author{Se-eun Yoon}
\affiliation{%
  \institution{University of California, San Diego}
  \city{La Jolla}
  \state{CA}
  \country{USA}}
\email{seeuny@ucsd.edu}

\author{Hyunsik Jeon}
\affiliation{%
  \institution{University of California, San Diego}
  \city{La Jolla}
  \state{CA}
  \country{USA}}
\email{hyjeon@ucsd.edu}

\author{Julian McAuley}
\affiliation{%
  \institution{University of California, San Diego}
  \city{La Jolla}
  \state{CA}
  \country{USA}}
\email{jmcauley@ucsd.edu}

\renewcommand{\shortauthors}{Yoon et al.}

\begin{abstract}
We introduce a multimodal dataset where users express preferences through images.
These images encompass a broad spectrum of visual expressions ranging from landscapes to artistic depictions.
Users request recommendations for books or music that evoke similar feelings to those captured in the images, and recommendations are endorsed by the community through upvotes.
This dataset supports two recommendation tasks: title generation and multiple-choice selection. 
Our experiments with large foundation models reveal their limitations in these tasks. 
Particularly, vision-language models show no significant advantage over language-only counterparts that use descriptions, which we hypothesize is due to underutilized visual capabilities.
To better harness these abilities, we propose the chain-of-imagery prompting, which results in notable improvements.
We release our code and datasets.
\end{abstract}

\ccsdesc[500]{Information systems~Information retrieval}
\keywords{conversational recommendation, vision-language models}



\maketitle

\begin{figure}[t]
    \centering
    \vspace{1em}
    \includegraphics[width=0.96\linewidth]{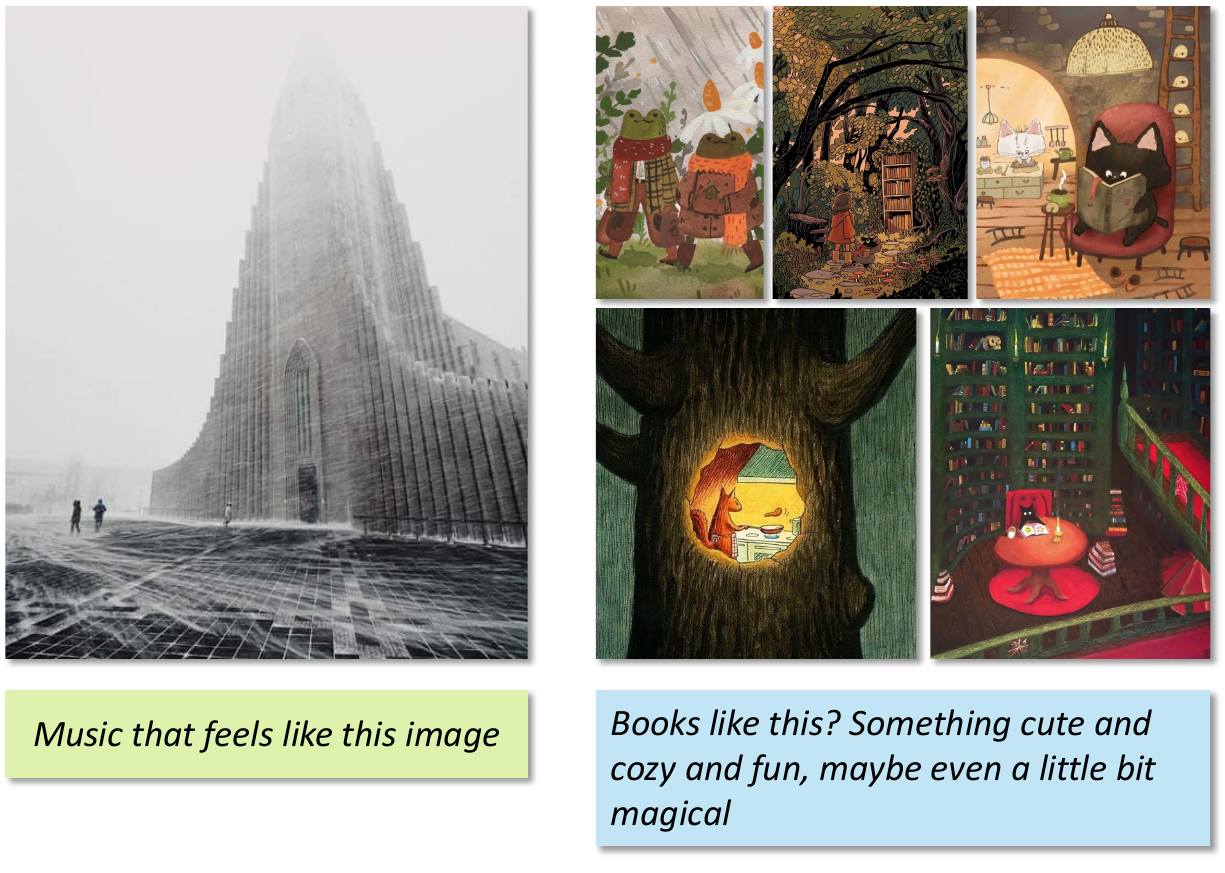}
    \caption{Users seeking recommendations through imagery. Each request uses a set of image(s) that conveys the essence of music (left) or book (right) one is looking for.}
    \label{fig:figure1}
\end{figure}

\section{Introduction}

Conversational recommender systems allow users to express preferences in natural language, yet this may not fully capture the diversity of user expression.
Consider the recommendation requests in Figure~\ref{fig:figure1}, found on Reddit.
One user wants music that `feels like this image', where the image evokes an ambiance that eludes easy description;
another user requests books that resonate with a set of images.
Such user behavior calls for models that can give recommendations based on multimodal expressions.

We present a novel benchmark to evaluate how well AI models understand and provide appropriate recommendations upon these visual-textual requests. 
We collect a dataset of recommendation requests as in Figure~\ref{fig:figure1}, each containing a set of images and supplementary text. 
These requests seek book or music recommendations using images spanning a wide range of styles and subjects. 
Each request is paired with a list of recommended items along with their respective upvote counts.
Based on this dataset, we construct two tasks that measure related but distinct recommendation capabilities: title generation (enlist relevant items) and multiple-choice selection (select the best item among a given candidate set).

These tasks are difficult due to the following challenges.
First, they require a nuanced understanding beyond the objects depicted, such as the mood, themes, and aesthetics.
Second, models should pair the visuals with books or music that evoke similar `feelings' that gather consensus among users.
Such pairing demands a capability beyond literal correspondence (e.g., linking the picture of a dog barking to the corresponding sound)~\cite{arandjelovic2018objects}.

Vision-language models (VLMs) are capable of understanding images, particularly through visually-conditioned language generation~\cite{liu2023llava, alayrac2022flamingo, achiam2023gpt}. 
On the other hand, large language models (LLMs) outperform conventional models in text-only conversational recommendation~\cite{he2023large}.
We seek to understand how zero-shot VLMs and LLMs perform in our multimodal conversational recommendation tasks.
For non-vision LLMs, we replace images with descriptions generated by VLMs.
Our experiments reveal that all models struggle, with the overall best model (GPT-4) achieving 67\% accuracy on a 5-way music selection task.
Further, VLMs do not offer much advantage over LLMs, possibly due to not fully utilizing their visual capabilities.
We propose a simple fix: to prompt VLMs to `look more closely,' and then to think of recommendations.
As for LLMs, only larger models benefit from more detailed descriptions.

Our contributions are threefold: (1) a multimodal dataset of real user requests and recommendations; (2) new benchmark tasks; (3) empirical results of six foundation models.
We release our code, datasets, and supplementary materials~\cite{supplementary}.

\section{Related Work}
\label{sec:related}

\boldheading{Multimodal Foundation Models.}
Multimodal foundation models are grouped into two categories: correspondence and generation.
Correspondence models focus on aligning different sensory modalities for tasks such as classification and retrieval, such as vision and text~\cite{RadfordKHRGASAM21, WangYYDT022} or audio and text~\cite{WuSKB22, HuangJLGLE22}. 
Generation models are better suited to open-ended tasks, such as visual question answering~\cite{wang2022ofa, xu2021vlm, alayrac2022flamingo, liu2023llava}.
Recent studies explore generation models for recommendation where item images can be flexibly incorporated as auxiliary input~\cite{wei2024unified, liu2024rec}.
Our work introduces generation models to a novel cross-modal scenario, where users make visual requests for non-visual creative content, necessitating a deeper understanding of user preferences from imagery.

\boldheading{Visual \& Conversational Recommendation.}
Visual information can enhance recommender systems, as in fashion~\cite{McAuleyTSH15, HeM16a, HeM16b} and art~\cite{HeFWM16}.
Conversations allow systems to interact directly with users through natural language~\cite{LiKSMCP18, ChenLZDCYT19, ZhouZBZWY20, LiXZ0Z022, RenTLR0XLRC22, WangZWZ22, he2023large}. 
Some studies explore visual conversational recommendation~\cite{guo2018dialog, saha2018towards, liao2021mmconv, du2023enhancing}.
Dialogues contain images of \textit{items}, focused on recognizing them or expressing preference regarding their visual aspects.
Our problem differs by presenting images devoid of items, challenging models to understand cross-modality, creativity, and emotional undertones.
While \citet{mckee2023language} explores audio retrieval for video clips and queries, each query is synthetically derived from its corresponding target audio.
Our dataset is from real users---it reflects their need to articulate requests through images instead of item descriptions.

\section{Problem Formulation}

\subsection{Datasets}

For data collection, we identify Reddit communities (\textit{subreddits}) that contain book and music recommendations: 
/r/booksthatfeellikethis (books), /r/musicsuggestions (music), /r/picturethesound (music), and /r/ifyoulikeblank (music). 
From these communities, we extract posts that request recommendations based on images, uploaded up to February 2024.
Such posts must contain at least one image and a text query asking for recommendations; we use the Python Reddit API Wrapper (PRAW)\footnote{https://praw.readthedocs.io/en/stable/} to identify these posts.
Particularly, the subreddits /r/musicsuggestions and /r/ifyoulikeblank contain posts other than recommendation requests, so we retrieve posts with text similar to the phrases `sound/feel like' or `fit' followed by `image', `picture', `photo', or `drawing';
/r/booksthatfeellikethis and /r/picturethesound consist mostly of recommendation requests, in accordance with the community guidelines.
After downloading all posts, we manually inspect each one to ensure that (1) the post is a recommendation request, and (2) the image cannot be trivially converted to text (e.g., screenshot of text or a list of items).
However, the invalid posts are few in proportion (2.9\%) to the entire data.
We denote the valid posts as \textit{requests}.

Next, we process the recommendations per request. 
Each request has comments from other users in a multi-level structure.
Typically, first-level comments directly provide recommendations by mentioning at least one item, and deeper-level comments are reactions to the recommendations.
We keep all the first-level comments and extract the items using gpt-3.5-turbo-0125.\footnote{https://platform.openai.com/docs/models/gpt-3-5-turbo} 
Human validation of 100 comments shows that no items are incorrectly extracted and 9\% are missed.
Prompt details are provided in our supplementary materials~\cite{supplementary}.
Additionally, in a given request $r$, each comment $c$ in the comment set $C_r$ is associated with a number $U(c)$ of upvotes,\footnote{This is the `net upvote,' which is the total number of upvotes minus the number of downvotes. We only consider positive net upvotes to ensure relevance.} which serves as an approximate measure of relevance. 
We assign the relevance score $S_r(i)$ for item $i$ as the sum of upvotes from all the comments that mentions the item: $S_r(i) = \sum_{c\in C_r} \mathbf{1}_{c \text{ mentions } i} \cdot U(c)$. 
We denote the set of recommended items for request $r$ as $I_r$.

We obtain 1,470 requests and 12,208 items for \textit{books}; 796 requests and 38,204 items for \textit{music}. 
Each request has an average of $3.11$ ($\pm3.14$) images and $14.04$ ($\pm12.55$) recommended items for \textit{books}; $1.41$ ($\pm1.74$) images and $65.92$ ($\pm101.00$) items for \textit{music}.
We report more data statistics in \cite{supplementary}.  
There is a notable upward trend in the number of requests in both domains, with the counts for the initial two months of 2024 surpassing those of any full previous year.

\begin{figure}[t]
    \centering
    \vspace{1em}
    \includegraphics[width=\linewidth]{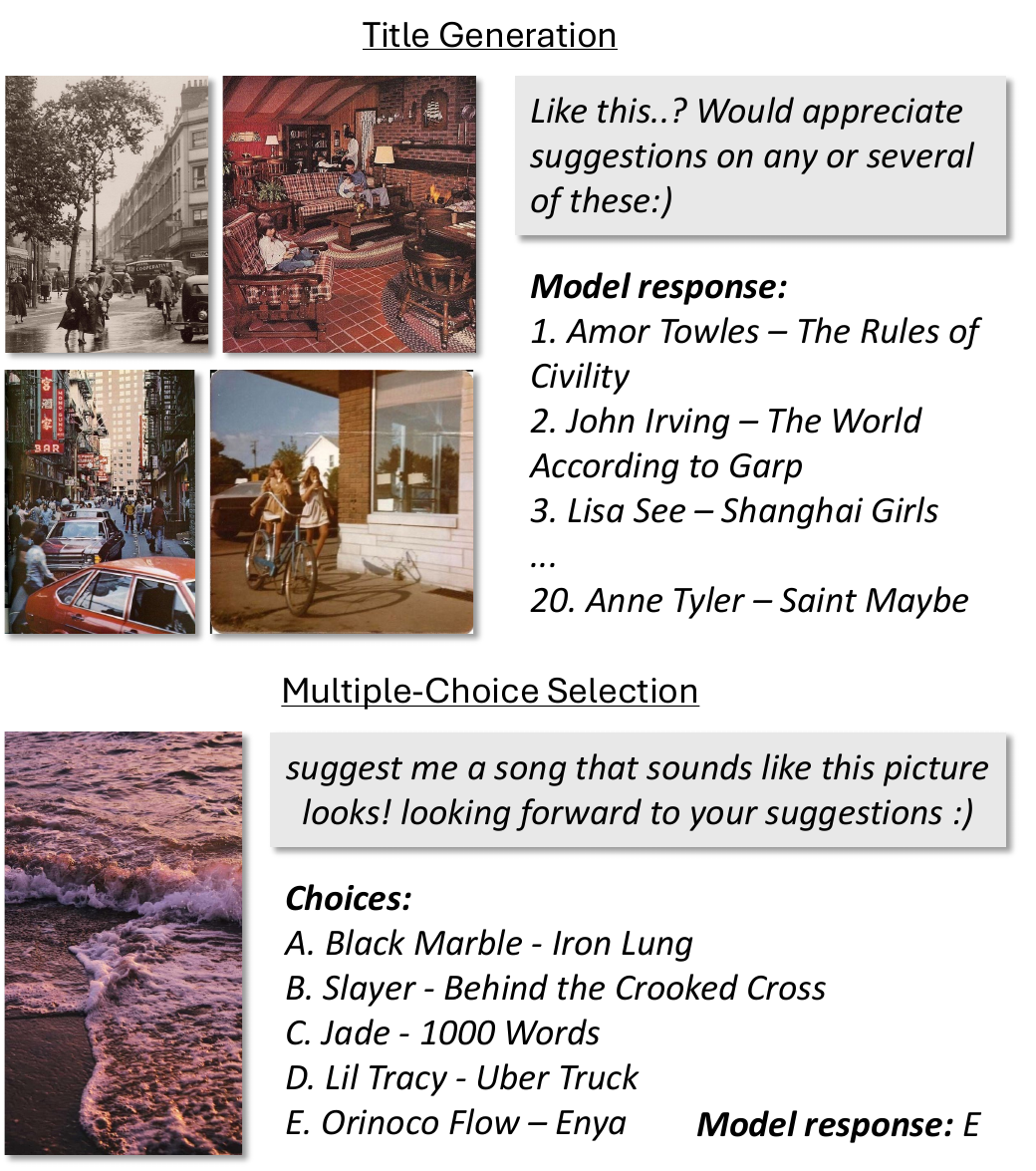}
    \caption{We construct title generation (above) and multiple-choice selection (below) tasks, based on our dataset containing requests and recommendations.}
    \label{fig:tasks}
\end{figure}

\subsection{Tasks}

We design two tasks to evaluate how AI models perform on multimodal conversational recommendation.
An overview is in Figure~\ref{fig:tasks}.

\boldheading{Title Generation.}
Given a request, we ask models to generate a list of item titles in the order of recommendation relevance. 
This approach follows the generative retrieval paradigm \cite{he2023large, de2020autoregressive}.
Since two or more items may have the same title, we specify models to output items in the \textit{Creator - Title} format, which we refer to as `title' for brevity.
We evaluate whether the top-$k$ generated titles contains any of the ground-truth recommendations.

\boldheading{Multiple-Choice Selection.}
Given a request and a candidate set of items, select the one that best matches the request.
This task evaluates a model's ability to \textit{select} the best item from those already listed, different from title generation, which measures its ability to \textit{retrieve}.
This separation also reflects the two-stage recommendation process, where candidate items are first retrieved and later re-ranked~\cite{liu2022neural, hron2021component}.\footnote{While we also considered a ranking task based on upvotes, we recognized that upvotes may be noisy, influenced by external factors such as timing.}
We generate candidate items as follows:
for each request $r$, the item with the highest score $i = \argmax_{i \in I_r} S_r(i)$ is designated as the positive item;
negative items are randomly sampled from the entire set of items, excluding the set of all relevant items $I_r$.
This candidate generation process minimizes potential noise, since it is very unlikely that a random item (not included in the recommendations) would be \textit{more suitable} than the item with the highest user consensus.
We set the number of negative items to $4$ and shuffle the candidate items to randomly assign options from A to E.\footnote{We empirically find that providing option tokens yields more properly formatted responses from models, compared to listing items without them.}
We compare the selected item with the ground-truth item.
\section{Methods}

We explore two vision-language models and four language-only models, all in zero-shot settings.
Since language-only models cannot take images as input, we first convert the images to descriptions using three different vision-language models.

\boldheading{Recommendation Models.}
For vision-language models, we use \textbf{GPT-4V} (gpt-4-1106-vision-preview)~\cite{achiam2023gpt} and \textbf{LLaVA} (llava-v1.5-13b)~\cite{liu2023llava}. 
GPT-4V is a proprietary model from OpenAI and LLaVA is an open-sourced model derived from pre-trained CLIP ViT-L/14~\cite{radford2021learning} and Vicuna~\cite{vicuna2023}.
For language-only models, we use \textbf{GPT-4} (gpt-4-0125-preview) and \textbf{GPT-3.5} (gpt-3.5-turbo-0125) from OpenAI; \textbf{LLaMA} (Llama-2-13b-chat-hf)~\cite{touvron2023llama} and \textbf{Vicuna} (vicuna-13b-v1.5)~\cite{vicuna2023}, both open-sourced.
Additionally, we explore GPT-4V with descriptions in place of images as input.\footnote{We were unable to do the same with LLaVA's code, since it requires an image input.}

\boldheading{Descriptions.}
We use either \textbf{OFA} (ofa-huge, 930M parameters)~\cite{wang2022ofa}, \textbf{LLaVA}, or \textbf{GPT-4V} to generate descriptions.
We instruct each model to \prompt{`Describe this image in detail, including its content, style, and vibe.'} 
For OFA, which struggles with customized instructions, we use the one which it was trained on: 
\prompt{`what does the image describe?'}
We find that each model progressively generates more detailed descriptions (OFA $<$ LLaVA $<$ GPT-4V), in terms of average description length (7, 95, 230 words per image, respectively) and human evaluation.\footnote{Upon testing 21 random images, an average of 5.86 out of 7 participants per image responded that GPT-4V gives a more detailed and accurate description than LLaVA.}

\begin{table}[t!]
\centering
\caption{Title generation (Hit@10) results with standard prompting. Vision models are marked with an asterisk (*). Results for $k={1, 20}$ show similar trends. (See \cite{supplementary}.)}
\label{table:generation}
\begin{tabularx}{\columnwidth}{llcccc}
\toprule
&  & \multicolumn{4}{c}{\textbf{Input modality}} \\
\cmidrule(lr){3-6}
 &  & \multicolumn{3}{c}{\textbf{Descriptions}} &  \\
\cmidrule(lr){3-5}
\textbf{Dataset} & \textbf{Model} & \textbf{OFA} & \textbf{LLaVA} & \textbf{GPT-4V} & \textbf{Raw images} \\
\midrule
Books   & Vicuna  & .0075 & 0 & 0 & . \\
        & LLaMA   & .1395 & .0918 & .0571 & . \\ 
        & LLaVA*   & . & . & . & .0456 \\ 
        & GPT-3.5 & .3503 & .4095 & .4510 & . \\
        & GPT-4   & .3837 & .4701 & \bf{.5048} & . \\ 
        & GPT-4V*  & .3823 & .4551 & .4830 & \underline{.4864} \\ 
\midrule
Music   & Vicuna  & .0251 & .0113 & .0013 & . \\
        & LLaMA   & .1068 & .1457 & .1344 & . \\ 
        & LLaVA*   & . & . & . & .1432 \\ 
        & GPT-3.5 & .1570 & .1922 & .2148 & . \\
        & GPT-4   & .1960 & .2688 & \bf .2990 & . \\
        & GPT-4V*  & .2023 & .2915 & .2927 & \underline{.2940} \\
\bottomrule
\end{tabularx}
\end{table}

\begin{table}[t!]
\centering
\caption{Title selection (\% accuracy) results with standard prompting.}
\label{table:selection}
\begin{tabularx}{\columnwidth}{llcccc}
\toprule
&  & \multicolumn{4}{c}{\textbf{Input modality}} \\
\cmidrule(lr){3-6}
 &  & \multicolumn{3}{c}{\textbf{Descriptions}} &  \\
\cmidrule(lr){3-5}
\textbf{Dataset} & \textbf{Model} & \textbf{OFA} & \textbf{LLaVA} & \textbf{GPT-4V} & \textbf{Raw images} \\
\midrule
Books   & Vicuna  & 28.98 & 35.65 & 34.69 & .\\
        & LLaMA   & 39.05 & 18.16 & 09.93 & .\\ 
        & LLaVA*   & . & . & . & 58.16 \\ 
        & GPT-3.5 & 60.61 & 67.07 & 68.10 & .\\
        & GPT-4   & 73.47 & 80.48 & \bf 82.18 & .\\ 
        & GPT-4V*  & 70.75 & 78.91 & 81.77 & \underline{80.82} \\ 
\midrule
Music   & Vicuna  & 26.51 & 35.93 & 32.79 & . \\
        & LLaMA   & 29.77 & 28.27 & 16.83 & . \\ 
        & LLaVA*   & . & . & . & 41.33 \\ 
        & GPT-3.5 & 45.73 & 60.68 & 59.30 & . \\
        & GPT-4   & 52.89 & 65.58 & \bf 66.96 & . \\
        & GPT-4V*  & 53.02 & 64.32 & \underline{66.45} & 64.70 \\
\bottomrule
\end{tabularx}
\end{table}

\boldheading{Prompts.}
We prompt models with an instruction, text query, and images (or descriptions).
An example of \textbf{standard prompting} for title generation is as follows: \prompt{`Given the request, provide recommendations. Enumerate 20 music pieces (1., 2., ...) in the order of relevance. Each piece should take the Artist - Title format. Don't say anything else.'}
We further ask if enhanced prompts improve performance.
Inspired by the success of chain-of-thought (CoT) prompting in logical reasoning tasks~\cite{wei2022chain}, we introduce \textbf{chain-of-imagery (CoI) prompting} to vision-language models. 
Unlike CoT, CoI prompting does not provide intermediate steps, since the process from images to books or music lacks clear, describable intermediates. 
Instead, CoI prompting simply adds extra guidelines to the standard instructions as follows:
\prompt{`Think step by step. First understand the given image(s) in detail, including its content, style, and vibe. Then, think of music pieces that capture the essence of the image(s).'}
We list all the prompts in \cite{supplementary}.

\boldheading{Evaluation Metrics.}
We use average Hit@$k$ ($k=\{1, 10, 20\}$) for title generation and accuracy for multiple-choice selection.

\begin{figure}[t!]
    \centering
    \vspace{1em}
    \includegraphics[width=0.85\linewidth]{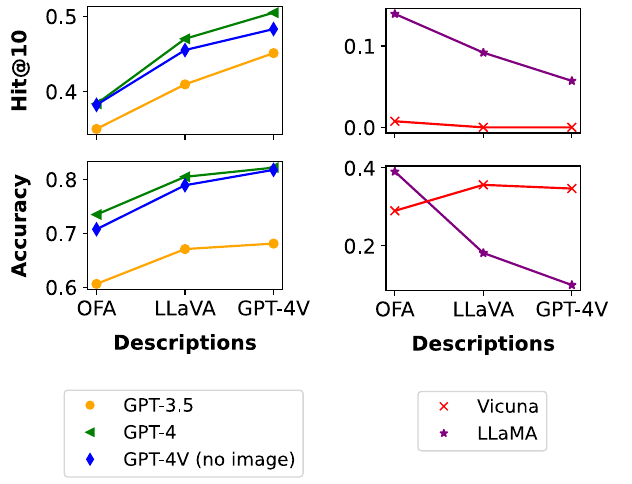}
    \caption{Larger models (left) benefit from detailed descriptions, but smaller models (right) do not. Results for \textit{books} are shown; results for \textit{music} are similar.}
    \label{fig:description_trend}
\end{figure}

\begin{table}[t!]
\centering
\caption{Chain-of-imagery (CoI) prompting improves GPT-4V by better leveraging vision capabilities, yielding better results compared to the text-only GPT-4 (from GPT-4V descriptions). Vision models (*) are given images as input.}
\label{table:coi-closed}
\begin{tabularx}{\columnwidth}{llcccc}
\toprule
\textbf{Dataset} & \textbf{Model} & \textbf{Hit@1} & \textbf{@10} & \textbf{@20} & \textbf{Acc (\%)} \\ 
\midrule
Books   & GPT-4V* + CoI & \bf .2136 & \bf .5299 & \bf .5986 & \bf 82.24 \\ 
        & GPT-4V*       & \underline{.2020} & .4864 & .5503 & 80.82 \\
        & GPT-4 (desc.)  & .1952 & \underline{.5048} & \underline{.5741} & \underline{82.18} \\
\midrule
Music   & GPT-4V* + CoI & \bf .1219 & \bf .3169 & \bf .3807 & \underline{65.45} \\
        & GPT-4V*       & \underline{.1181} & .2940 & \underline{.3568} & 64.70 \\
        & GPT-4 (desc.)  & .1005 & \underline{.2990} & .3354 & \bf{66.96} \\
\bottomrule
\end{tabularx}
\end{table}

\begin{table}[t!]
\centering
\caption{LLaVA's improvement with chain-of-imagery prompting is limited to title selection (accuracy).}
\label{table:coi-open}
\begin{tabularx}{\columnwidth}{llcccc}
\toprule
\textbf{Dataset} & \textbf{Model} & \textbf{Hit@1} & \textbf{@10} & \textbf{@20} & \textbf{Acc (\%)} \\ 
\midrule
Books   & LLaVA* + CoI & .0150 & .0361 & .0381 & \bf 61.56 \\
        & LLaVA*       & \bf .0218 & \bf .0456 & \bf .0497 & 58.16 \\
\midrule
Music   & LLaVA* + CoI & .0540 & .1206 & .1545 & \bf 45.00 \\
        & LLaVA* & \bf .0678 & \bf  .1432 & \bf .1759 & 41.33 \\
\bottomrule
\end{tabularx}
\end{table}

\section{Results}

We run the models on our task and draw four takeaway messages.

\boldheading{[M1] Models struggle, particularly smaller ones.} 
Tables~\ref{table:generation} and \ref{table:selection} show results for standard prompting.
The best performing model is GPT-4 using descriptions generated by GPT-4V (we discuss later why GPT-4V, a vision model, does not perform as well as text-only GPT-4).
For title generation (Table~\ref{table:generation}), GPT-4 is able to include any of the ground-truth items in its top-10 list less than than 51\% (\textit{books}) and 30\% (\textit{music}) of the time.
For selection (Table~\ref{table:selection}), its accuracy is 82.18\% (\textit{books}) and 66.96\% (\textit{music}).
While GPT-4V closely follows and GPT-3.5 shows a moderate margin for improvement, smaller models (LLaVA, LLaMA, Vicuna) perform significantly worse.
In some cases, the model fails to retrieve any ground truth items (e.g., Vicuna at Table~\ref{table:generation}), or makes choices worse than random selection (e.g., LLaMA at Table~\ref{table:selection}). 

\boldheading{[M2] Only larger models benefit from detailed descriptions.}
Figure~\ref{fig:description_trend} shows performance trends as descriptions increase in detail (OFA $\rightarrow$ LLaVA $\rightarrow$ GPT-4V).
Larger models (all of which we use are closed-sourced) show improved performance as descriptions become more detailed, although the gains are diminishing. 
Smaller models do not follow this trend; models often perform worse as detail increases.
Particularly, descriptions that span multiple phrases (LLaVA) can sometimes be more helpful than single-phrase descriptions (OFA), but making descriptions even more detailed (GPT-4V) is consistently detrimental.
One reason may be that smaller (13B) models are poor at handling long inputs~\cite{liu2024lost}.

\boldheading{[M3] Using descriptions can be better than using images.} 
Language-only models cannot `see' the image and instead rely on descriptions. 
Descriptions can provide important details but may not always accurately or fully represent the image itself.
However, we observe that descriptions are sometimes sufficient, or even slightly better, to be used in place of images.
Tables~\ref{table:generation} and \ref{table:selection} show that, compared to GPT-4V using images, GPT-4 achieves similar performance with LLaVA descriptions and slightly higher performance with GPT-4V descriptions.
In the case of open-sourced models, LLaMA consistently outperforms LLaVA in title generation for \textit{books}, although model size is similar (13B).
This observation raises an important question: \textit{How can we better harness the visual capabilities of VLMs?}

\boldheading{[M4] Chain-of-imagery (CoI) prompting may help VLMs better harness their visual capabilities.}
CoI prompting asks VLMs to first understand the images in detail, and then to think of relevant items that match the images. 
Table~\ref{table:coi-closed} shows that this strategy improves GPT-4V, making GPT-4V + CoI the overall best performing approach.
We also try CoI prompting for text-only GPT-4, but it is not as effective (see~\cite{supplementary}).
This observation implies that there is indeed more to be `seen' from the images than descriptions alone, which can be utilized by nudging a VLM.
However, while CoI prompting helps LLaVA in title selection, it negatively impacts title generation (Table~\ref{table:coi-open}).
We hypothesize that CoI prompting benefits performance that is already above a certain level;
LLaVA lags significantly behind GPT-4V in title generation, but not as much in selection.
Testing with more VLMs is required to confirm this hypothesis, which we suggest as future research.

\section{Conclusion and Limitations}

In this work, we deliver a new dataset for multimodal conversational recommendation and propose two benchmark tasks, title generation and multiple-choice selection.
We test VLMs and LLMs and provide four takeaway messages on model performance, input modalities, and prompting strategies.
While we provide initial insights, further experiments are necessary for a deeper understanding of model behaviors. 

\bibliographystyle{ACM-Reference-Format}
\bibliography{references}

\clearpage



\end{document}